\documentclass[runningheads]{llncs}

\usepackage{eccv}
\usepackage{eccvabbrv}
\usepackage{graphicx}
\usepackage{booktabs}
\usepackage[accsupp]{axessibility}
\usepackage{hyperref}
\usepackage{orcidlink}

\usepackage[table]{xcolor}
\usepackage{setspace}
\usepackage{amssymb}   
\usepackage{wasysym}   
\usepackage{fontawesome5}

\begin{document}

\title{Uni-World VLA: Interleaved World Modeling and Planning for Autonomous Driving} 

\titlerunning{Uni-World VLA}

\author{
Qiqi Liu\inst{1,2,3}$^{\S,\ast}$ \and
Huan Xu\inst{3}$^{\ast}$ \and
Jingyu Li\inst{1,2,3}$^{\S}$ \and
Bin Sun\inst{3}$^{\dagger}$ \and
Zhihui Hao\inst{3}$^{\dagger}$ \and
Dangen She\inst{3} \and
Xiatian Zhu\inst{4} \and
Li Zhang\inst{1,2}$^{\ddagger}$
}

\authorrunning{Q. Liu et al.}

\institute{
Fudan University
\and
Shanghai Innovation Institute
\and
Li Auto Inc.
\and
University of Surrey
}

\maketitle

\begin{center}
\faGithub\
\textcolor{cyan}{
\href{https://github.com/LogosRoboticsGroup/UniWorldVLA}{
\texttt{github.com/LogosRoboticsGroup/UniWorldVLA}
}}
\end{center}

\let\thefootnote\relax
\footnotetext[1]{
$^{\ast}$ Equal contribution; 
$^{\dagger}$ Project Leader; 
$^{\ddagger}$ Corresponding author; 
$^{\S}$ Intern at Li Auto Inc.
}

\begin{abstract}
Autonomous driving requires reasoning about how the environment evolves and planning actions accordingly. 
Existing world-model-based approaches typically predict future scenes first and plan afterwards, resulting in open-loop imagination that may drift from the actual decision process. 
In this paper, we present \textbf{Uni-World VLA}, a unified vision-language-action (VLA) model that tightly interleaves future frame prediction and trajectory planning. 
Instead of generating a full world rollout before planning, our model alternates between predicting future frames and ego actions step by step, allowing planning decisions to be continuously conditioned on the imagined future observations. 
This interleaved generation forms a closed-loop interaction between world modeling and control, enabling more adaptive decision-making in dynamic traffic scenarios. 
In addition, we incorporate monocular depth information into frames to provide stronger geometric cues for world modeling, improving long-horizon scene prediction. 
Experiments on the NAVSIM benchmark show that our approach achieves competitive closed-loop planning performance while producing high-fidelity future frame predictions. 
These results demonstrate that tightly coupling world prediction and planning is a promising direction for scalable VLA driving systems.
  \keywords{Autonomous driving \and World modeling and planning \and Depth fusion}
\end{abstract}

\section{Introduction}
\label{sec:intro}

\begin{figure}[tb]
  \centering
  \includegraphics[width=\linewidth]{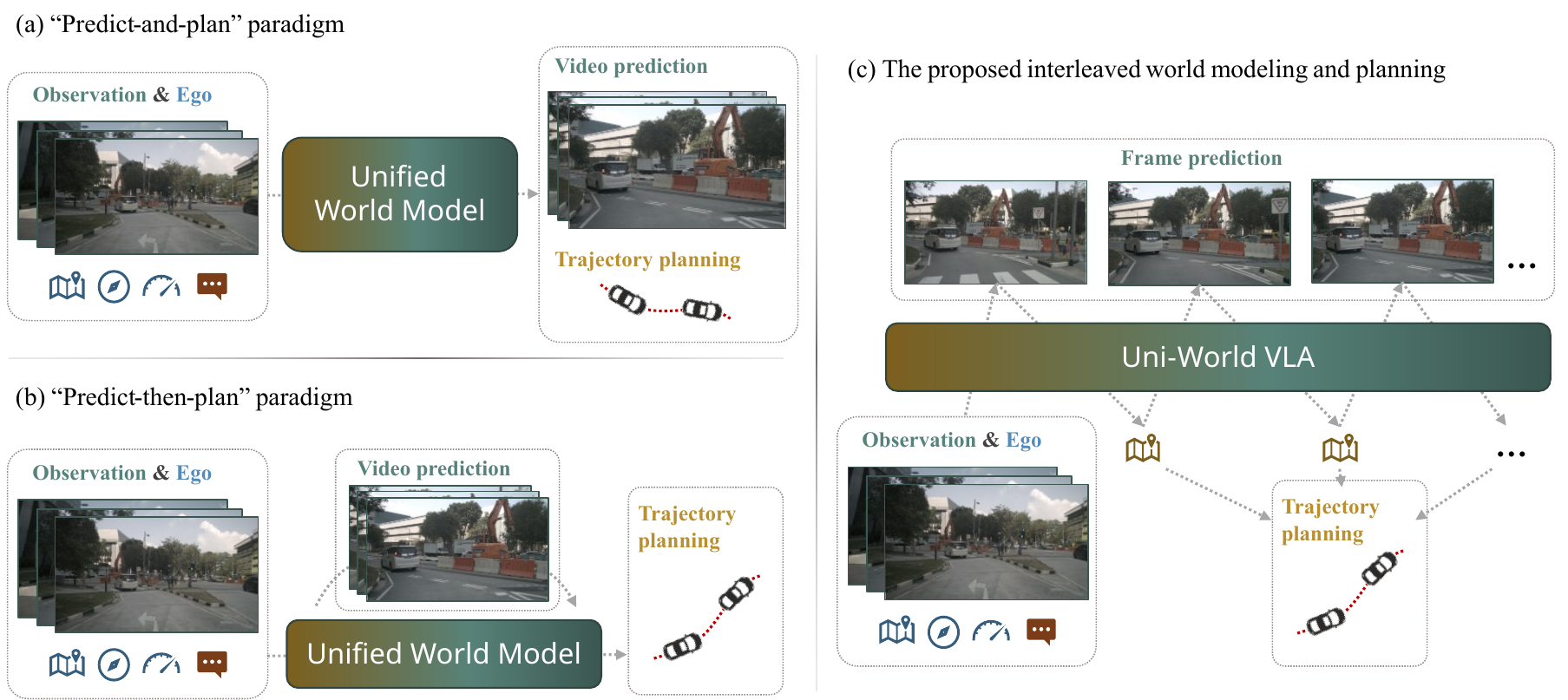}
  \caption{Different generative paradigms of unified world models for autonomous driving. (a) Unified world models perform video generation and planning as separate tasks; (b) World-conditioned trajectory prediction, where future trajectories are predicted conditioned on the generated world states; (c) Interleaved world modeling and planning (ours). Visual tokens and action queries are generated alternately, forming a closed-loop interaction that respects the temporal causality of driving.
  }
  \label{fig:comparison}
\end{figure}

With the rapid advancement of multi-modal large models~(MLLM) and generative models, Vision-Language-Action~(VLA) models~\cite{senna,zhou2025autovla,li2026sgdrive} and world models~\cite{vista,zhang2025epona} have attracted increasing attention in autonomous driving. Unlike conventional end-to-end approaches~\cite{zhang2025perception,zhangfuture,liao2025diffusiondrive} that primarily rely on imitation learning, VLA-based methods~\cite{li2025recogdrive,zhou2025autovla} leverage rich semantic understanding to predict the ego vehicle’s future trajectory, while driving world models~\cite{chen2024drivinggpt,zheng2024doe} forecast future scene evolution based on learned physical dynamics. Although these approaches have achieved impressive performance, they are typically developed in isolation, addressing trajectory prediction and environment modeling separately, and thus fail to effectively share and integrate complementary knowledge between the two paradigms.

To enable knowledge sharing between world modeling and trajectory prediction, prior works~\cite{li2025imagidrive,zhao2025pwm,zhang2025epona,bartoccioni2025vavim} have combined both objectives in a unified framework. Based on the temporal ordering of the tasks, these methods fall into two paradigms: parallel ``predict-and-plan'' and sequential ``predict-then-plan''. In the ``predict-and-plan'' paradigm~\cite{chen2024drivinggpt,bartoccioni2025vavim}~(Fig.~\ref{fig:comparison}~(a)), world modeling and planning are integrated within a single autoregressive architecture. Despite joint training, the tasks remain functionally decoupled: world modeling focuses on high-fidelity next-frame prediction conditioned on actions, while trajectory planning maps visual observations to control outputs without explicitly leveraging the learned dynamics. In contrast, the ``predict-then-plan'' paradigm~\cite{zhao2025pwm,zhang2025epona,li2025imagidrive}~(Fig.~\ref{fig:comparison}~(b)) first predicts future scenes and then generates the ego vehicle’s trajectory conditioned on them. A key limitation is the implicit assumption that the environment remains stationary, whereas real-world traffic is inherently non-stationary, with continuous interactions between the ego agent and surrounding vehicles.

Regardless of whether the ``predict-and-plan'' or ``predict-then-plan'' paradigm is adopted, both suffer from a common limitation. In complex urban scenarios (e.g., unprotected left turns or merging), traffic conditions evolve rapidly. If a world model generates a multi-second rollout~(e.g., 4 seconds) conditioned on an initial intent, it effectively produces a ``frozen hallucination'' of the future. This rollout assumes that the environment will react to a fixed plan that is not updated by the observations generated during prediction. Consequently, when the planner relies on later portions of the rollout~(e.g., the third second), the visual evidence may already be outdated, as it does not reflect subtle ego-vehicle adjustments made earlier in the horizon~(e.g., slight braking or steering at 0.5 seconds).
To address this limitation, we propose an interactive prediction-planning framework that tightly couples world modeling and planning within a single model, shown in Fig.~\ref{fig:comparison}~(c). Instead of separating prediction and planning, our approach performs future scene prediction and trajectory planning in an interleaved manner. At each step, the model first predicts the future world state conditioned on historical observations and past trajectories, and then plans the ego-vehicle action based on the predicted state. This step-wise interaction enables the planner to continuously update its decisions according to the evolving environment, forming an interleaved prediction-planning process.

Specifically, we design a unified autoregressive architecture that alternates between generating visual tokens and action tokens. Visual observations are first compressed into a discrete token space for efficient sequence modeling, and an MLLM~\cite{xie2024showo} autoregressively predicts an interleaved sequence of future scene and ego action tokens. In this formulation, scene tokens represent the evolving world state and action tokens correspond to the ego-vehicle trajectory. Such alternating generation establishes an interaction between world modeling and planning, allowing decisions to be continuously refined based on newly predicted observations and reducing error accumulation over long horizons. To further improve temporal sensitivity, we follow \cite{zhao2025pwm} to introduce a bi-directional intra-frame attention mechanism that enables tokens within the same frame to capture rich spatial dependencies while preserving causal masking across time. In addition, to provide complementary geometric cues, monocular depth maps are estimated using Depth Anything 3~\cite{lin2025depth3recoveringvisual}, and the resulting depth features are fused with historical frames through a cross-attention mechanism, which improves the quality of future frame prediction.

Our contributions are summarized as follows:
{\bf (i)} We introduce \textbf{interleaved modeling and planning}, a step-wise feedback paradigm that tightly couples world modeling and trajectory planning. To realize this idea, we develop a unified autoregressive architecture that alternately generates future visual tokens and action tokens, forming an interaction between prediction and control and enabling planning decisions to be continuously refined based on newly predicted observations.
{\bf(ii)} We further introduce a \textbf{depth integration} strategy that incorporates monocular depth maps and fuses the geometric features with historical frames through cross-attention, providing complementary spatial cues for future frame prediction.
{\bf (iii)} We evaluate the proposed approach on the high-fidelity driving benchmark NAVSIM. Extensive experiments show that the interleaved generation mechanism improves both planning performance and video prediction quality.

\section{Related Work}
\label{sec:relatedwork}

\noindent\textbf{VLA models for autonomous driving.}
Fully autonomous driving remains a central goal in AI and robotics~\cite{chib2024e2eadsurvey,huThu2025visionlanguageactionmodelsautonomous}. Early Vision-Action models map raw sensory inputs to control commands or trajectory waypoints via imitation or reinforcement learning~\cite{chib2024e2eadsurvey,le2022survey,liang2018cirl}, but often struggle with high-level reasoning, interpretability, and flexible decision-making~\cite{Grigorescu2020survey, hu2023planningorientedautonomousdriving}. VLA models integrate scene understanding, language-grounded reasoning, and actionable outputs~\cite{kim2024openvlaopensourcevisionlanguageactionmodel,huThu2025visionlanguageactionmodelsautonomous}. Language-mediated approaches reformulate planning as language generation: DriveGPT4~\cite{xu2024drivegpt4interpretableendtoendautonomous}, GPT-Driver~\cite{mao2023gptdriverlearningdrivegpt}, DriveMLM~\cite{wang2023drivemlm}, DriveLM~\cite{sima2023drivelm}, and AutoVLA~\cite{zhou2025autovla} explore chain-of-thought reasoning, structured tokenization, masked language modeling, graph-based visual QA, and unified reasoning-action frameworks, respectively. Dual-system VLAs decouple high-level deliberation from trajectory planning, e.g., DriveVLM~\cite{tian2024drivevlm}, DiffVLA~\cite{jiang2025diffvla}, and VLP~\cite{pan2024vlp}. Most prior methods, however, operate open-loop and lack iterative modeling of future states. Our approach addresses this by interleaving predicted future frames and actions within a unified prompt, enabling closed-loop visual-action feedback that improves planning robustness.

\noindent\textbf{3D scene reconstruction.}
Accurate 3D scene representation is critical for autonomous driving~\cite{deng2026best3dscenerepresentation,chib2024e2eadsurvey}, providing geometric priors for tasks like object detection, path planning, and collision avoidance~\cite{zheng2024gaussianad,chen2025geodrive}. Early approaches rely on geometric pipelines such as structure-from-motion~\cite{Schonberger2016sfmrevisited, pan2025globalsfmrevisited} and multi-view stereo~\cite{Seitz2006mvs,pepe2022uavplatformsandthesfm-mvsapproach}. Neural rendering methods, including implicit radiance fields~\cite{mildenhall2021nerf,xiao2025neuralradiancefieldsreal} and explicit 3D Gaussian Splatting~\cite{kerbl20233dgaussian,huang2024textits3gaussian}, enable high-quality, scalable scene reconstruction. Feed-forward geometry estimation directly predicts scene structure~\cite{chirstodoulides2025surveyon3drecon}, while large-scale depth foundation models—Depth Anything 3~\cite{lin2025depth3recoveringvisual}, MapAnything~\cite{keetha2026mapanything}, VGGT~\cite{Wang2025vggt}, and $\pi^3$~\cite{wang2025pi3}—provide generalizable priors, joint depth-structure prediction, and robust multi-view inference. Together, these advances equip autonomous driving systems with long-horizon spatial understanding~\cite{li2025comprehensivesurveyworldmodels}.

\noindent\textbf{World models for autonomous driving.}
World models learn internal predictive representations to simulate future states from actions~\cite{li2025comprehensivesurveyworldmodels}, relying on robust 3D and spatiotemporal scene cues~\cite{chen2025geodrive,gao2025rad,yuan2024presight}. Recent methods move beyond visual forecasting toward interactive, policy-aware simulation: PWM~\cite{zhao2025pwm} jointly predicts states and actions; UniDrive-WM~\cite{xiong2026unidrivewm} couples scene understanding, planning, and generative modeling; ImagiDrive~\cite{li2025imagidrive} integrates VLM-driven imagination; SGDrive~\cite{li2026sgdrive} adds hierarchical scene-to-goal reasoning; Infinite-World~\cite{wu2026infiniteworld} scales to thousand-frame horizons via hierarchical memory; and ResWorld~\cite{zhang2026resworld} models dynamic agents with temporal residuals. Most prior methods rely solely on RGB inputs, limiting geometric reasoning; we introduce depth-informed conditioning in historical visual prompts to enhance spatial awareness without explicit depth modeling in future-frame generation.

\section{Methods}
\label{sec:methods}

\begin{figure}[tb]
  \centering
  \includegraphics[width=0.8\linewidth]{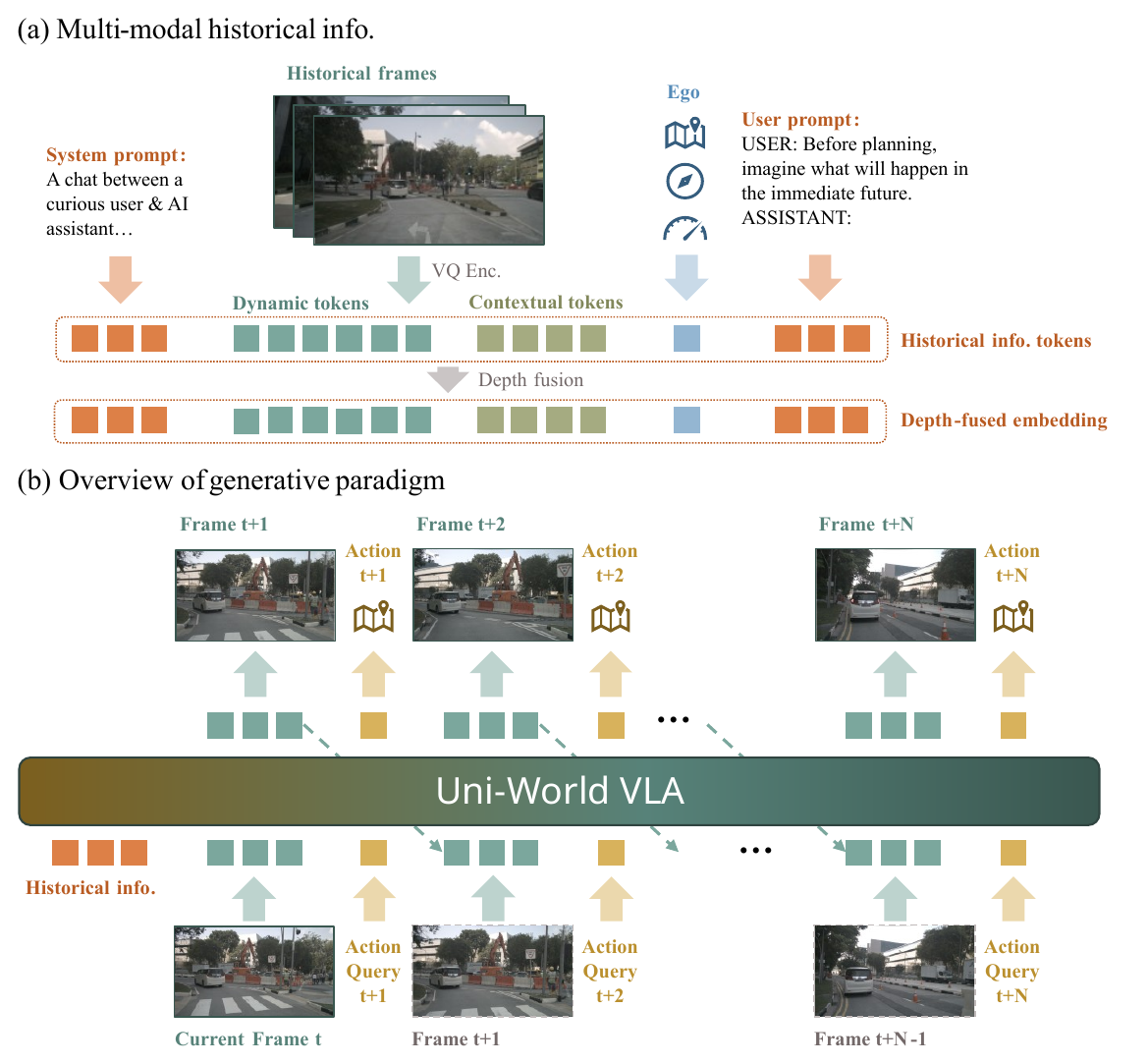}
  \caption{Overview of the paradigm of alternative generation in Uni-World VLA. (a) The construction of multi-model historical information; (b) The interleaved frame-action generative paradigm.  }
  \label{fig:overview}
\end{figure}

\noindent\textbf{Overview.}
Fig.~\ref{fig:overview} illustrates the overall paradigm of our framework for future-frame and trajectory prediction. Given past ego-centric frames and auxiliary state information, visual observations are first encoded into discrete tokens using a MagVIT-v2~\cite{yu2024magvitv2}. Depth cues estimated by Depth Anything 3~\cite{lin2025depth3recoveringvisual} are then fused with historical visual tokens via cross attention to provide complementary geometric information. These historical inputs representing ego-state and vision, are fed into Show-o~\cite{xie2024showo}, a Phi-1.5-based~\cite{Li2023phi-1.5} multimodal LLM. The LLM then autoregressively generates an interleaved sequence of future frame and action tokens, which are finally decoded into RGB frames by the MagVIT-v2 decoder and trajectories by an MLP.

\noindent\textbf{Inputs and tokenization.}
Let \(\{I_{t-M}, \dots, I_{t-1}\}\) denote the \(M\) historical ego-centric image frames (\(t\) denotes the current time, \(I \in \mathbb{R}^{H \times W \times 3}\), where $H$ and $W$ are the image height and width). To capture both environmental context and temporal dynamics, we partition the historical video stream in NAVSIM~\cite{daniel2024navsim,cao2025pseudosimulation} into two complementary modalities. \texttt{Contextual tokens} correspond to the initial high-resolution frames that provide detailed scene semantics and structural information, while \texttt{dynamic tokens} are sampled at 10\,Hz and at lower resolution to capture fine-grained motion cues and short-term temporal variations.

In addition to image data, we collect auxiliary ego status at time \(t\), which includes the velocity, acceleration, and the high-level driving command of the ego vehicle.
Each raw image \(I\) is encoded into a sequence of discrete visual tokens 
\begin{equation}
{c, d} = \mathrm{Encoder}_{\mathrm{MagVIT}}(I)
\end{equation}
using the MagVIT-v2, where \(c\) denotes the contextual tokens and \(d\) denotes the dynamic tokens. This tokenization produces a compact symbolic representation that is convenient for autoregressive sequence modeling by the LLM backbone. Moreover, the system and user text prompts are also encoded, enabling the LLM to better comprehend the scenario and the underlying task.
As shown in Fig.~\ref{fig:overview}(b), the model input is constructed as a short chat-style context:
\texttt{[System Prompt | Dynamic \& Contextual Tokens | User Prompt | Ego Tokens]},
where the \texttt{System Prompt} defines the assistant’s general behavior, encouraging helpful and detailed responses within a conversational setting  and the \texttt{User Prompt} specifies the task objective, explicitly requiring the model to anticipate the immediate future before planning. \texttt{Ego tokens} concatenate ego velocity, ego acceleration, and the high-level driving command at time $t$, which represents the current dynamic state and navigation intent. The LLM consumes the mixed token stream and performs autoregressive generation in an interleaved manner. More details of the unified tokenizer can be found in Appendix~\ref{appendix:tokens}.

\noindent\textbf{Interleaved frame-action generation.}
In our implementation, we generate up to \(N=8\) future frames. With a temporal interval of 0.5 seconds per frame, this corresponds to a 4.0-second prediction horizon. 
Unlike purely sequential video prediction, we adopt an alternative generative paradigm with step-wise interaction between visual prediction and action querying. After generating each future frame \(t+k\), an action query corresponding to the same timestamp is fed back to the LLM to infer the predicted ego position at that time step. The predicted state is then incorporated into the subsequent generation process, forming an interleaved prediction loop (as shown in Fig.~\ref{fig:overview} (b)):
\begin{align}
\hat d_{t+k} &\sim p_\theta(d_{t+k} \mid \hat d_{\le t+k-1}, \hat a_{\le t+k-1}), \\
\hat a_{t+k} &\sim p_\theta(a_{t+k} \mid \hat d_{\le t+k}, \hat a_{\le t+k-1}),
\end{align}
where \(\hat d\) and \(\hat a\) respectively represent the dynamic tokens and the action tokens.

\noindent\textbf{Decoding and output.}
Each predicted discrete token sequence $\hat d_{t+k}$ is decoded by the MagVIT-v2 decoder into the corresponding RGB frame, conditioned on the contextual token $c_{t+2\lfloor k/2 \rfloor}$ that provides visual guidance at the per-second scale,
\begin{equation}
\hat I_{t+k} = \mathrm{Decoder}_{\mathrm{MagVIT}}(\hat d_{t+k}; c_{t+2\lfloor k/2 \rfloor}).
\end{equation}
The final visual output thus forms a sequence of reconstructed future frames 
\(\{\hat I_{t+1}, \dots, \hat I_{t+N}\}\) sampled at 0.5-second intervals. 
In parallel, the predicted action tokens are passed through an MLP head to obtain the corresponding ego positions 
\(\hat a_{t+1}, \hat a_{t+2}, \dots, \hat a_{t+N}\), 
which together constitute the planned trajectory over a 4.0-second horizon.

\begin{figure}[tb]
  \centering
  \includegraphics[width=\linewidth]{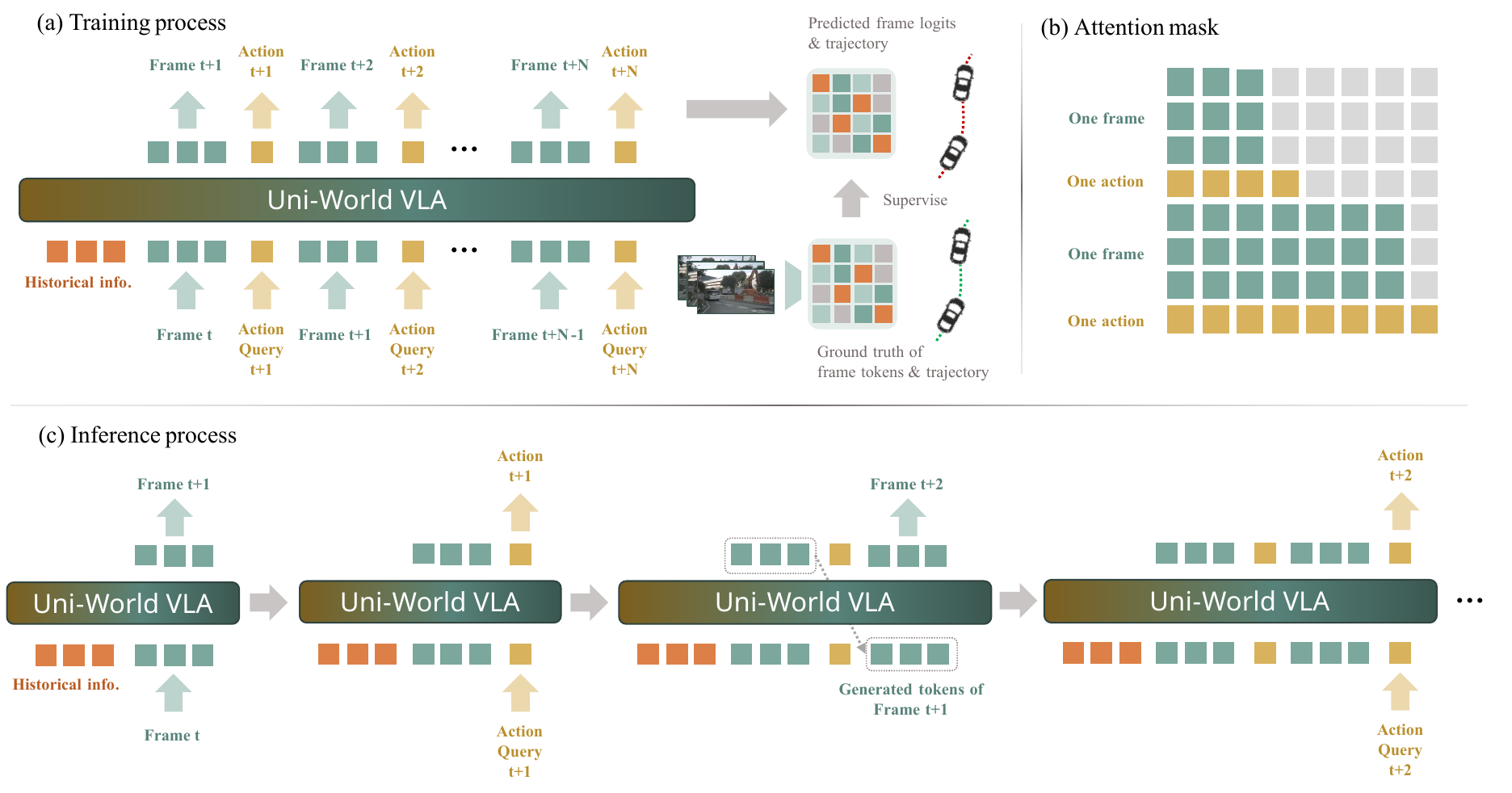}
  \caption{Schematic illustration of training and inference process.
  (a) Interleaved sequence for joint video generation and trajectory supervision.
  (b) Causal attention mask.
  (c) Autoregressive interleaved inference with KV-cache reuse.}
  \label{fig:TrainInfProcess}
\end{figure}

\noindent\textbf{Training objectives.}
The model is trained with joint supervision on both future visual token generation and trajectory prediction. During the training phase, future frames and action queries are arranged in an interleaved order as shown in the Fig.~\ref{fig:TrainInfProcess} (a), and are processed together by the LLM for both inference and supervision.

For visual prediction, the model outputs logits over the discrete MagVIT-v2~\cite{yu2024magvitv2} vocabulary for each future frame token. \cite{zhao2025pwm} points out, if simply supervise the logits with cross-entropy loss, a significant proportion of the frame tokens tend to keep unchanged across adjacent frames. To alleviate this issue, wo adopt the Dynamic Focal Loss~\cite{zhao2025pwm} that emphasize temporally varying image regions through spatial weighting. Specifically, we define the dynamic weight \(\omega(d_{t+k}^i, d_{t+k-1}^i)\) as:
\begin{equation}
\omega(d_{t+k}^i, d_{t+k-1}^i)=\alpha \mathbb{I}(d_{t+k}^i\ne d_{t+k-1}^i)+\beta \mathbb{I}(d_{t+k}^i=d_{t+k-1}^i), \quad \alpha > \beta
\end{equation}
where \(\mathbb{I}(\cdot)\) is the indicator function and \(\alpha\), \(\beta\) are hyperparameters that control the relative importance of dynamic versus static token predictions. The overall dynamic-weighted cross-entropy is defined as:
\begin{equation}
\mathcal{L}_{\text{dyn}} 
= -\frac{1}{N}\sum_{k=1}^{N}\sum_{i=1}^{L} 
\omega(d_{t+k}^i, d_{t+k-1}^i)\log p_\theta(d_{t+k}^i \mid \hat d_{<t+k}^i, \hat a_{<t+k}^i),
\end{equation}
where \(L\) denotes the total number of predicted visual tokens within one frame.

For trajectory prediction, the hidden states corresponding to action tokens are fed into an MLP head to regress the ego positions at each future step. 
We supervise the predicted trajectory \(\hat{a}_{t+1:t+N}\) using an L1 loss:
\begin{equation}
\mathcal{L}_{\text{traj}} 
= \frac{1}{N} \sum_{k=1}^{N} 
\left\| \hat{a}_{t+k} - a_{t+k} \right\|_1.
\end{equation}

The final training objective is a weighted sum of the two terms:
\begin{equation}
\mathcal{L} = \lambda_1 \mathcal{L}_{\text{dyn}} + \lambda_2 \mathcal{L}_{\text{traj}},
\end{equation}
where \(\lambda_1\), \(\lambda_2\) balance visual generation and trajectory supervision.

The corresponding attention mask is illustrated in Fig.~\ref{fig:TrainInfProcess} (b). During generation of each future frame, newly generated visual tokens can attend to all previous tokens as well as all tokens within the current frame. This design allows the model to capture both temporal dependencies across past frames and spatial dependencies within a frame, facilitating coherent prediction of adjacent visual regions. Historical frames are processed with the same attention mask construction, ensuring consistent intra-frame and inter-frame context throughout the training sequence.

\noindent\textbf{Inference.}
At inference, future frames and actions are generated step by step in an autoregressive manner (Fig.~\ref{fig:TrainInfProcess} (c)). Starting from the current frame \(I_t\), the model first generates the visual tokens for \(t+1\), which are then decoded into the corresponding RGB frame. An action query for \(t+1\) is subsequently fed to the LLM to predict the ego action at the same timestamp. The generated visual tokens for \(t+1\) are appended to the context to condition the generation of \(t+2\), and this process repeats until all \(N\) future frames and actions are produced. The attention masking scheme mirrors training. To improve efficiency, we leverage a KV-cache: the key and value representations from previous steps are stored and reused, so that the LLM only computes attention for newly generated tokens instead of reprocessing the entire sequence. This interleaved generation loop allows the model to iteratively reason about scene dynamics and ego-motion throughout the prediction horizon.

\noindent\textbf{Depth integration.}
We adopt Depth Anything 3~\cite{lin2025depth3recoveringvisual}, a state-of-the-art monocular depth estimation model, to extract depth maps from input images. The depth map extraction process is expressed as:
\begin{equation}
D = \text{DepthAnything3}(I),
\end{equation}
where $I \in \mathbb{R}^{H \times W \times 3}$ represents the input image and $D \in \mathbb{R}^{H \times W}$ is the extracted depth map.

To match input images, the extracted depth map $D$ is resized into two different resolutions: 256×448 and 128×224. The two resized depth maps are then fed into two improved VIT (base on MagVIT-v2~\cite{yu2024magvitv2}) modules, which are named context-depth-encoder (CDE) and dynamic-depth-encoder (DDE) correspondingly and serve as depth encoders.

The input image $I$ is first subjected to vector quantization to generate image feature tokens, which are further divided into context tokens and dynamic tokens corresponding to the two resolutions of the depth map. These context tokens and dynamic tokens are then embedded separately to generate context token embeddings and dynamic token embeddings (as shown in Fig.~\ref{fig:overview} (a)), which serve as queries in the cross-attention mechanism. Specifically, the context token embeddings are used as queries to fuse with depth feature embeddings output by the CDE, and the dynamic token embeddings are used as queries to fuse with depth feature embeddings output by the DDE:
\begin{equation}
E_{q,c} = \text{Embed}(c), E_{q,d} = \text{Embed}(d),
\end{equation}
where $E_{q,c} \in \mathbb{R}^{N_{q,c} \times d}$ denotes the context token embeddings (embedded results of context tokens), and $E_{q,d} \in \mathbb{R}^{N_{q,d} \times d}$ denotes the dynamic token embeddings (embedded results of dynamic tokens).
The visual token embeddings $E_{q,c}$ and $E_{q,d}$ act as queries and are fused with the corresponding keys and values from CDE and DDE, respectively. The cross-attention (CA) fusion processes are formulated as:
\begin{equation}
E_{\text{fused},c} = \text{CA}(E_{q,c}, D_{k,c}, D_{v,c}),
E_{\text{fused},d} = \text{CA}(E_{q,d}, D_{k,d}, D_{v,d}).
\end{equation}
In the above formulas, $D_{k,c}$ (key) and $D_{v,c}$ (value) are generated by the CDE, corresponding to the context query $E_{q,c}$; $D_{k,d}$ and $D_{v,d}$ are generated by the DDE, corresponding to the dynamic query $E_{q,d}$. Then, the fused feature embeddings $E_{\text{fused},c}$ and $E_{\text{fused},d}$ are fed into the VLM model. 

\section{Experiments}
\label{sec:experiments}

\subsection{Experimental Setup}

\noindent\textbf{Dataset.} 
We conduct all experiments on the \textbf{NAVSIM}~\cite{daniel2024navsim, cao2025pseudosimulation} driving dataset, a high-fidelity simulated benchmark providing ego-centric RGB sequences, vehicle state information, and structured annotations for planning evaluation. We follow the official train/validation/test split and adopt a fixed temporal interval of 0.5\,s between frames. Unless otherwise specified, the model predicts \(N=8\) future frames (4.0 seconds horizon).
For the ablation study, some variants require 10 Hz sampled trajectories, while NAVSIM only provides 2 Hz logs. Therefore, the full 10 Hz trajectories are supplemented from \textbf{nuPlan}~\cite{caesar2021nuplan}.

\noindent\textbf{Evaluation metrics.} 
\label{sec:metrics}
We evaluate our method using both planning-oriented and video generation metrics. Planning performance is measured by the Predictive Driver Model Score (\textbf{PDMS})~\cite{daniel2024navsim}, which includes five sub-metrics: no-at-fault collisions~(NC), drivable area compliance~(DAC), time-to-collision~(TTC), comfort~(Comf.), and ego progress~(EP). Video prediction quality is evaluated using Fréchet Video Distance~(\textbf{FVD})~\cite{unterthiner2018fvd}, which measures distribution-level realism of generated videos.

\noindent\textbf{Implementation details.}
Our model is initialized from Policy World Model (PWM)~\cite{zhao2025pwm}, 
which itself is fine-tuned from Show-o~\cite{xie2024showo}. 
For visual tokenization, we adopt the compressive MagVIT-v2~\cite{yu2024magvitv2} tokenizer 
pretrained under the dual-branch framework of PWM. 
The tokenizer consists of two branches with separate codebooks of size $8192$ each. 
The frozen high-resolution branch processes input images of resolution 
$256 \times 448$ and produces $448$ contextual tokens per frame. 
The pretrained low-resolution branch processes images of resolution 
$128 \times 224$ and generates $28$ dynamic tokens per frame. In addition, the weights of the depth encoders are initialized using the weights of MagVIT-v2.
For Uni-World VLA, we fine-tune the model on NAVSIM for $30$ epochs, 
selecting the best checkpoint based on PDMS (achieved at epoch $16$).

The deep fusion training adopted in this paper follows a two-stage progressive paradigm, which is designed to balance the stability of feature extraction and the adaptability of cross-module fusion, ensuring the model converges stably and achieves effective feature fusion. 

We use AdamW as the optimizer and train the model on $32$ NVIDIA H20 GPUs. 
Only the front-view camera is used as input, with a training batch size of $3$. 
The model takes $2$ seconds of historical observations to autoregressively predict 
$8$ future frames and corresponding waypoints over a $4$-second horizon, where the learning rate follows a cosine annealing schedule.
The specific training process and key parameter configurations are detailed as follows.

\begin{spacing}{1.2}
\textbf{Stage 1: Pre-training of Deep Feature Extraction Module}
\end{spacing}

The CDE and DDE modules are trained in this stage. We adopt an action-free video prediction setup to extract depth information; specifically, we only generate future frames at 10 Hz within 1 second for supervision. To preserve the representation capability of the pre-trained foundation model, its weights are completely frozen, while all other trainable parameters are unfrozen for targeted optimization. The training is conducted with a total of 5 epochs, and the learning rate is set to \(3 \times 10^{-5}\) to ensure stable parameter update and initial convergence of the module.

\begin{spacing}{1.2}
\textbf{Stage 2: Multi-modal Joint Training} 
\end{spacing}

Building upon the first stage, the second stage focuses on joint optimization. The pre-trained CDE and DDE modules are frozen to maintain their effective deep feature extraction capability, while the fusion module and the foundation model are unfrozen to promote their collaborative learning. We adopt \textbf{Scheme E} (Sec.~\ref{sec:scheme-E}) to supervise future frames and trajectories. This stage is trained for 16 epochs, and the learning rate is set to \(2 \times 10^{-5}\), aiming to realize accurate coupling of features and full convergence of the overall model.

\subsection{Main Results}

\begin{table}[h]
\centering
\small
\setlength{\tabcolsep}{4pt}
\caption{\textbf{Comparison with state-of-the-art methods on the NAVSIM test split.} 
PDMS and its sub-metrics (NC, DAC, EP, TTC, and Comfort; see Sec.~\ref{sec:metrics}) evaluate closed-loop driving performance. 
Input modalities include multi-view cameras (C), single-view camera (SC), and multi-view cameras with LiDAR (C\&L). 
Best results are shown in \textbf{bold}.}
\begin{tabular}{l|c|ccccc|>{\columncolor{gray!20}}c}
\toprule
Method & Input & NC $\uparrow$ & DAC $\uparrow$ & EP $\uparrow$ & TTC $\uparrow$ & Comf.\,$\uparrow$ & PDMS $\uparrow$ \\
\midrule
\multicolumn{8}{l}{\textit{Traditional End-to-End Methods}} \\
VADv2-$\nu_{8192}$~\cite{chen2024vadv2} & C & 97.2 & 89.1 & 76.0 & 91.6 & \textbf{100.0} & 80.9 \\
UniAD~\cite{hu2023planningorientedautonomousdriving} & C & 97.8 & 91.9 & 78.8 & 92.9 & \textbf{100.0} & 83.4 \\
TransFuser~\cite{chitta2023transfuser} & C\&L & 97.7 & 92.8 & 79.2 & 92.8 & \textbf{100.0} & 84.0 \\
ReCogDrive-IL~\cite{li2025recogdrive} & SC & 98.1 & 94.7 & 80.9 & 94.2 & \textbf{100.0} & 86.5\\
DiffusionDrive~\cite{liao2025diffusiondrive} & C\&L & 98.2 & 96.2 & 82.2 & 94.7 & \textbf{100.0} & 88.1 \\
\midrule
\multicolumn{8}{l}{\textit{World Model Methods}} \\
DrivingGPT~\cite{chen2024drivinggpt} & SC & \textbf{98.9} & 90.7 & 79.7 & 94.9 & 95.6 & 82.4 \\
Epona~\cite{zhang2025epona} & SC & 97.9 & 95.1 & 80.4 & 93.8 & 99.9 & 86.2 \\
ImagiDrive-A~\cite{li2025imagidrive} & SC & 98.1 & 96.2 & 80.1 & 94.4 & \textbf{100.0} & 86.9 \\
DriveVLA-W0~\cite{li2025drivevla} & SC & 98.4 & 95.3 & 80.9 & 95.4 & \textbf{100.0} & 87.2 \\
SGDrive-IL~\cite{li2026sgdrive} & SC & 98.6 & 95.1 & 81.2 & 95.4 & \textbf{100.0} & 87.4 \\
PWM~\cite{zhao2025pwm} & SC & 98.6 & 95.9 & 81.8 & 95.4 & \textbf{100.0} & 88.1 \\
WoTE~\cite{li2025wote} & C\&L & 98.5 & \textbf{96.8} & 81.9 & 94.9 & 99.9 & 88.3 \\
ResWorld~\cite{zhang2026resworld} & C\&L & \textbf{98.9} & 96.5 & 83.1 & 95.6 & \textbf{100.0} & 89.0 \\
Uni-World VLA (Ours) & SC & 98.7 & 96.7 & \textbf{83.2} & \textbf{96.1} & \textbf{100.0} & \textbf{89.4} \\
\bottomrule
\end{tabular}
\label{tab:navsim_pdms}
\end{table}

Table~\ref{tab:navsim_pdms} reports closed-loop planning performance on the NAVSIM test split. Uni-World VLA achieves the highest PDMS (89.4), indicating a strong balance of safety, comfort and forward progress, and attains the best EP (83.2) and TTC (96.4), suggesting improved efficiency of forward progress and execution safety. ResWorld~\cite{zhang2026resworld} records marginally higher NC but uses multi-sensor input (camera + LiDAR); its aggregate PDMS (89.0) remains slightly below ours, highlighting the effectiveness of our single-view camera-only design. Other baselines (e.g., DrivingDPT~\cite{chen2024drivinggpt}, WoTE~\cite{li2025wote}) perform well on individual metrics but do not match our combined PDMS. 

\begin{table}[ht]
\centering
\small
\setlength{\tabcolsep}{4pt}
\caption{\textbf{Comparison of video generation quality across driving world models.} 
FVD~\ref{sec:metrics} measures the realism of generated future video sequences. 
We additionally report the maximum prediction horizon, frame rate, dataset, and camera view used by each method.}
\resizebox{\linewidth}{!}{
\begin{tabular}{l|cccccc}
\toprule
Metric \& Settings & WoVoGen~\cite{lu2024wovogen} & DriveDreamer~\cite{wang2023drivedreamer} & SVD~\cite{blattmann2023stablevideodiffusion,chen2024drivinggpt} & DrivingGPT~\cite{chen2024drivinggpt} & GenAD~\cite{yang2024genadgeneralizedpredictivemodel} & Ours \\
\midrule
\rowcolor{gray!15}
FVD$\downarrow$ & 417.7 & 340.8 & 227.5 & 142.6 & 184.0 & \textbf{141.8} \\
Max Duration/Fps & 2.5\,s/2\,Hz & 4\,s/2\,Hz & 4\,s/2\,Hz & 4\,s/2\,Hz & 4\,s/2\,Hz & 4\,s/2\,Hz \\
Dataset & nuScenes & nuScenes & NAVSIM & NAVSIM & OpenDV & NAVSIM \\
View & Multi & Multi & Front & Front & Front & Front \\
\bottomrule
\end{tabular}}
\label{tab:video_metrics}
\end{table}

\begin{figure}[ht]
  \centering
  \includegraphics[width=\linewidth]{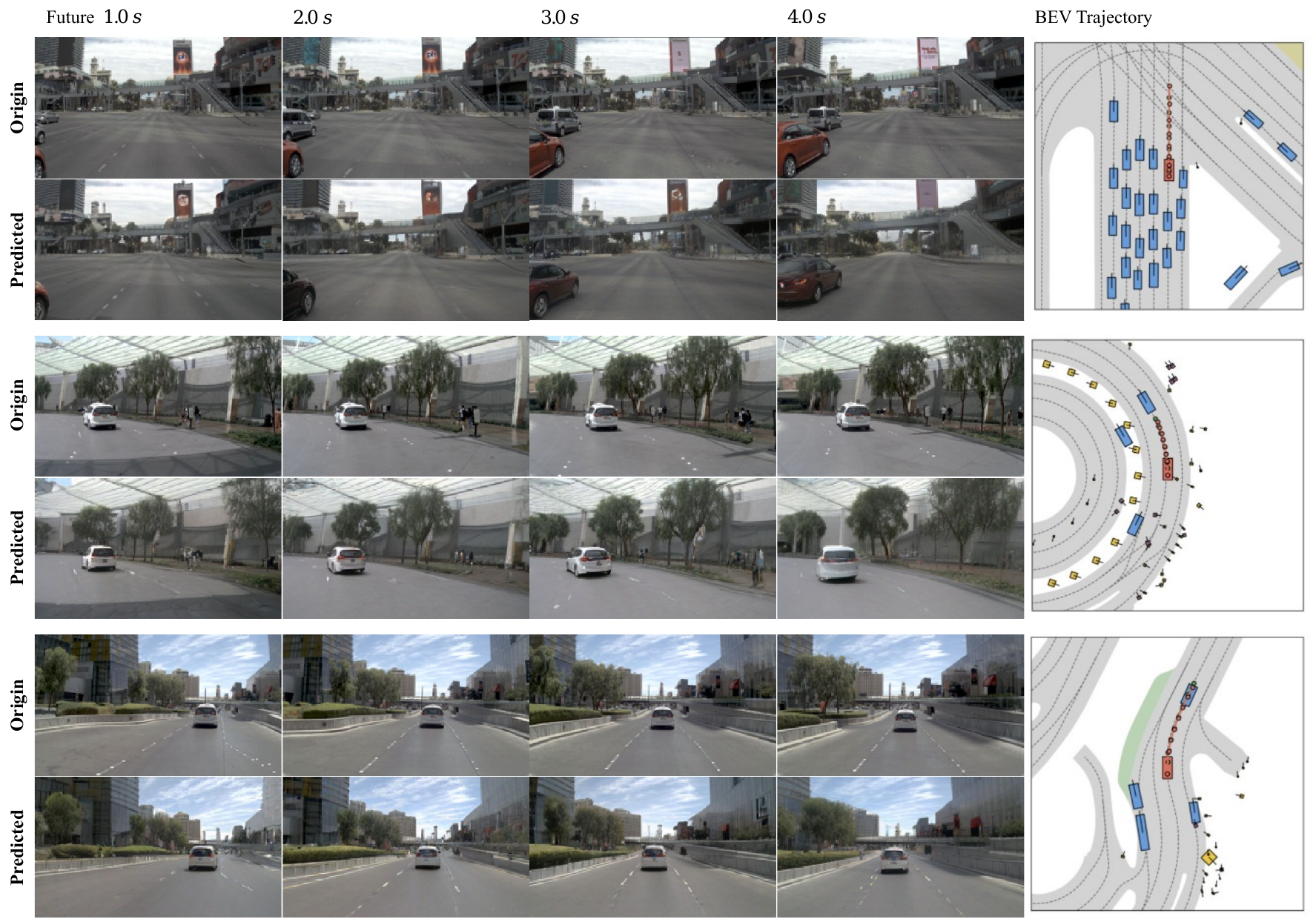}
  \caption{Visualization of predicted frames and BEV trajectories}
  \label{fig:Visualization}
\end{figure}

Table~\ref{tab:video_metrics} compares 2-Hz short-horizon video generation quality on NAVSIM and related benchmarks using FVD. 
Our method achieves a competitive FVD of 141.8 under the 4\,s/2\,Hz NAVSIM protocol, 
outperforming SVD~\cite{blattmann2023stablevideodiffusion,chen2024drivinggpt}, GenAD~\cite{yang2024genadgeneralizedpredictivemodel}, DrivingGPT~\cite{chen2024drivinggpt}, and several prior generative baselines.
More importantly, when considered together with the closed-loop results in Table~\ref{tab:navsim_pdms}, our model delivers substantially stronger planning performance (PDMS 89.4 vs.\ 82.4 of DrivingGPT, Table~\ref{tab:navsim_pdms}). 
This demonstrates that Uni-World VLA maintains competitive visual generation quality while achieving superior planning reliability and safety. 

Overall, Uni-World VLA maintains high comfort and drivable-area compliance while enhancing forward progress and TTC-based safety through the proposed interleaved generative planning scheme. Our approach also exhibits stronger holistic competitiveness by effectively balancing video fidelity and closed-loop planning performance.

Fig.~\ref{fig:Visualization} provides qualitative comparisons of predicted future frames and corresponding BEV trajectories. 
Our model produces temporally consistent visual dynamics while generating smooth and safe planning trajectories that respect lane geometry and surrounding agents. 
Compared to prior generative baselines, the predicted motions are more stable and better aligned with feasible driving behaviors, further supporting the quantitative gains in PDMS and TTC.

\subsection{Ablation Study}

\noindent\textbf{Effect of pretrain, future frames and depth.}
We ablate three design choices (Table~\ref{tab:ablation_future}): using a pretrained checkpoint, generating future frames to inform trajectory planning, and incorporating image depth as an additional conditioning signal.

\begin{table}[h]
\centering
\small
\setlength{\tabcolsep}{4pt}
\caption{Ablation study on the effects of pretraining, future-frame modeling, and depth conditioning. NC, DAC, EP, TTC, Comfort, and PDMS measure planning quality, while FVD evaluates the quality of generated future frames.}
\resizebox{\linewidth}{!}{
\begin{tabular}{ccc|ccccc|>{\columncolor{gray!20}}c|>{\columncolor{gray!15}}c}
\toprule
Pretrain & Future Frames & Depth 
& NC $\uparrow$ 
& DAC $\uparrow$ 
& EP $\uparrow$ 
& TTC $\uparrow$ 
& Comf.\ $\uparrow$ 
& PDMS $\uparrow$ 
& FVD $\downarrow$ \\
\midrule
$\times$ & $\times$ & $\times$ 
& 97.1 & 91.4 & 77.4 & 91.5 & \textbf{100.0} & 82.1 & -\\
$\checkmark$ & $\times$ & $\times$ 
& \textbf{98.8} & 95.8 & 82.0 & 95.8 & \textbf{100.0} & 88.2 & -\\
$\checkmark$ & $\checkmark$ & $\times$ 
& \textbf{98.8} & 96.5 & 82.9 & \textbf{96.4} & \textbf{100.0} & 89.2 & 164.2\\
$\checkmark$ & $\checkmark$ & $\checkmark$ 
& 98.7 & \textbf{96.7} & \textbf{83.2} & 96.1 & \textbf{100.0} & \textbf{89.4} & \textbf{141.8}\\
\bottomrule
\end{tabular}
}
\label{tab:ablation_future}
\end{table}

\begin{figure}[htb]
  \centering
  \includegraphics[width=0.9\linewidth]{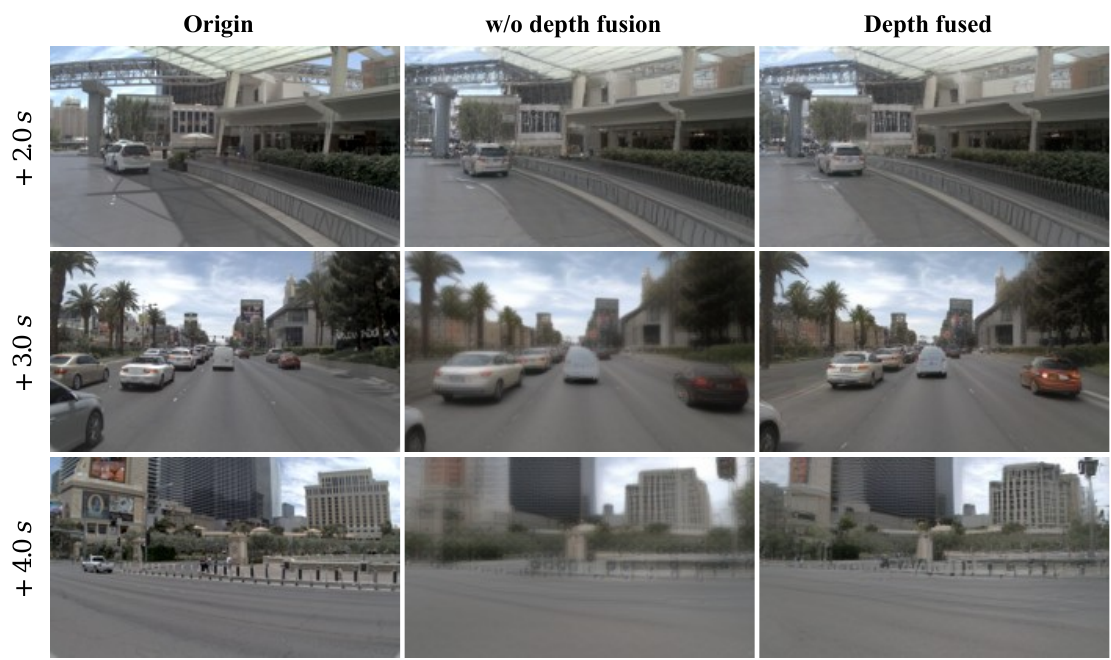}
  \caption{Comparison of predicted future frames with and without depth fusion}
  \label{fig:DepthAblaVis}
\end{figure}

Pretraining provides the largest improvement, increasing PDMS from 82.1 to 88.2 its subscores.
Enabling future-frame generation further improves performance, raising PDMS from 88.2 to 89.2 with consistent gains in DAC, EP, and TTC. This suggests explicitly modeling future world states provides additional context for trajectory planning.
Adding depth information further boosts performance: DAC increases to 96.7, EP reaches 83.2 and video quality improves, with FVD dropping from 164.2 to 141.8. Overall, combining pretraining, future-frame modeling, and depth conditioning yields the best results, showing these components complement each other for world modeling and planning.

Fig.~\ref{fig:DepthAblaVis} compares future-frame predictions with and without depth fusion. At 2.0\,s, both reconstruct the scene reasonably well, but by 3.0\,s, the model without depth shows blurred structures in fast-driving scenarios. At 4.0\,s, particularly in turns, depth fusion preserves clearer spatial layouts and geometric cues, keeping predictions smooth and coherent despite the coarse 2-Hz sampling.

\noindent\textbf{Effect of different alternative generation schemes.}
Table~\ref{tab:ablation_scheme} compares five alternative frame-action generation schemes without depth integration under the same NAVSIM evaluation protocol. The detailed token orders are illustrated by the formulas below.

\begin{table}[ht]
\centering
\caption{Ablation over five alternative generation schemes (without depth fusion).}
\label{tab:ablation_scheme}
\begin{tabular}{l|ccccc|>{\columncolor{gray!20}}c}
\toprule
Scheme
& NC $\uparrow$ 
& DAC $\uparrow$ 
& EP $\uparrow$ 
& TTC $\uparrow$ 
& Comf.\ $\uparrow$ 
& PDMS $\uparrow$ \\
\midrule
A-cross-frequency alternation & \textbf{98.8} & 96.4 & 81.2 & 96.0 & \textbf{100.0} & 88.3 \\
B-high-frequency actions-frames & 98.7 & 95.1 & 77.3 & \textbf{96.4} & \textbf{100.0} & 86.1\\
C-hybrid dense-then-coarse & 98.7 & 95.8 & 80.6 & 96.0 & \textbf{100.0} & 87.8 \\
D-sliding 1s action windows & 98.6 & 94.5 & 78.4 & 95.2 & 99.7 & 85.7 \\
E-2Hz aligned interleaving(Ours) & \textbf{98.8} & \textbf{96.5} & \textbf{82.9} & \textbf{96.4} & \textbf{100.0} & \textbf{89.2} \\
\bottomrule
\end{tabular}
\end{table}

\textbf{Scheme A} introduces a simple cross-frequency interleaving between frames and actions. Despite the frequency mismatch, it slightly improves over the PWM baseline (PDMS 88.3), suggesting that early action conditioning on partial visual context can be beneficial.
\textbf{Scheme B} enforces strict 10\,Hz frame-action alternation and dense action prediction. This leads to a noticeable performance drop (PDMS 86.1), likely due to the mismatch between dense training supervision and the 2\,Hz evaluation protocol.
\textbf{Scheme C} adopts a hybrid strategy with dense alternation in the first second followed by coarse 2\,Hz action generation. This partially mitigates the mismatch and improves performance over Scheme B (PDMS 87.8).
\textbf{Scheme D} predicts sliding 1\,s action windows for each frame. The overlapping horizons introduce conflicting supervision signals, resulting in the lowest PDMS (85.7).
\label{sec:scheme-E}
\textbf{Scheme E} aligns frame and action generation directly at the 2\,Hz evaluation frequency with strict F→A (frame to action) alternation. This configuration achieves the best overall performance (PDMS 89.2), indicating that matching generation frequency with the planning/evaluation protocol improves temporal consistency and planning quality.

\begin{center}
\resizebox{\linewidth}{!}{
\begin{tabular}{ll}
\textbf{A:} & \texttt{[Prompt]→[0.1s F]→[0.5s A]→[0.2s F]→[1.0s A]→...→[0.8s F]→[4.0s A]→[0.9s F]→[1.0s F]} \\

\textbf{B:} & \texttt{[Prompt]→[0.1s F]→[0.1s A]→[0.2s F]→[0.2s A]→...→[1.0s F]→[1.0s,1.1s,...,3.9s,4.0s A]} \\

\textbf{C:} & \texttt{[Prompt]→[0.1s F]→[0.1s A]→[0.2s F]→[0.2s A]→...→[1.0s F]→[1.0s,1.5s,2.0s,...,4.0s A]} \\

\textbf{D:} & \texttt{[Prompt]→[0.1s F]→[0.1s 0.2s ... 1.0s A]→[0.2s F]→[0.2s 0.3s ... 1.1s A]→...→[0.9s F]→}\\
&\texttt{[0.9s 1.0s ... 1.8s A]→[1.0s F]→[0.1s 0.2s ... 1.9s 2.0s 2.5s ... 4.0s A]} \\

\textbf{E:} & \texttt{[Prompt]→[0.5s F]→[0.5s A]→[1.0s F]→[1.0s A]→...→[4.0s F]→[4.0s A]} \\
\end{tabular}
}
\end{center}

\noindent\textbf{Effect of historical visual information.}
Table~\ref{tab:ablation_history} studies the effect of historical visual information. 
Using both contextual and dynamic tokens over a 2.0\,s history yields the best overall performance, achieving the highest PDMS (89.2), EP (82.9), and the lowest FVD (164.2), indicating stronger closed-loop planning and generation quality. 
Reducing the history to 1.0\,s slightly improves NC and TTC but degrades PDMS and FVD, suggesting a trade-off between conservative safety margins and overall performance. 
Using only contextual tokens maintains competitive PDMS and the best DAC, highlighting the importance of high-resolution scene structure, whereas using only dynamic tokens significantly degrades both planning and generation metrics. 
Overall, contextual tokens provide spatial semantics, dynamic tokens provide motion cues, and their combination over a longer history offers the best balance.

\begin{table}[h]
\centering
\caption{Effect of historical visual information on planning (without depth fusion).}
\setlength{\tabcolsep}{0.5pt}
\label{tab:ablation_history}
\begin{tabular}{l|ccccc|>{\columncolor{gray!20}}c|>{\columncolor{gray!20}}c}
\toprule
Historical Visual Info.
& NC $\uparrow$ 
& DAC $\uparrow$ 
& EP $\uparrow$ 
& TTC $\uparrow$ 
& Comf.\ $\uparrow$ 
& PDMS $\uparrow$ 
& FVD $\downarrow$ \\
\midrule
2.0\,s Context+Dynamic (Ours) 
& 98.8 & 96.5 & \textbf{82.9} & 96.4 & \textbf{100.0} & \textbf{89.2} & \textbf{164.2} \\
1.0\,s Context+Dynamic 
& \textbf{99.0} & 96.4 & 81.4 & \textbf{96.7} & \textbf{100.0} & 88.8 & 170.7 \\
Context Only 
& 98.6 & \textbf{96.8} & 82.3 & 96.2 & \textbf{100.0} & 89.1 & 165.5 \\
Dynamic Only 
& 97.4 & 90.8 & 76.2 & 92.3 & \textbf{100.0} & 81.7 & 203.6 \\
\bottomrule
\end{tabular}
\end{table}

\section{Conclusion}
We have introduced Uni-World VLA, a unified VLA model that interleaves world prediction and trajectory planning. In this model, a unified autoregressive architecture alternates between future frame prediction and trajectory planning, creating a tight feedback loop between predicted environment evolution and control. We also proposed enhancements to bolster this framework: the incorporation of monocular depth features via cross-attention. In experiments on the NAVSIM high-fidelity simulator, Uni-World VLA significantly outperforms prior state-of-the-art methods. It attains the highest overall driving score (PDMS) and improves both the time-to-collision (TTC) safety metric and ego progress (EP), while maintaining competitive video prediction quality (FVD) compared to baseline world models. These results demonstrate that the interleaved prediction-and-planning strategy with depth fusion enables more robust, context-aware decision-making in complex driving scenarios.

\newpage

\bibliographystyle{splncs04}
\bibliography{main}

@String(CVPR  = {IEEE Conf. Comput. Vis. Pattern Recog.})

@String(ECCV  = {Eur. Conf. Comput. Vis.})

@String(NeurIPS = {Adv. Neural Inform. Process. Syst.})

@String(CVPR  = {CVPR})

@String(ECCV  = {ECCV})

@String(NeurIPS = {NeurIPS})

@misc{kim2024openvlaopensourcevisionlanguageactionmodel,
      title={OpenVLA: An Open-Source Vision-Language-Action Model}, 
      author={Moo Jin Kim and Karl Pertsch and Siddharth Karamcheti and Ted Xiao and Ashwin Balakrishna and Suraj Nair and Rafael Rafailov and Ethan Foster and Grace Lam and Pannag Sanketi and Quan Vuong and Thomas Kollar and Benjamin Burchfiel and Russ Tedrake and Dorsa Sadigh and Sergey Levine and Percy Liang and Chelsea Finn},
      year={2024},
      eprint={2406.09246},
      archivePrefix={arXiv},
      primaryClass={cs.RO},
}

@misc{huThu2025visionlanguageactionmodelsautonomous,
      title={Vision-Language-Action Models for Autonomous Driving: Past, Present, and Future},
      author={Tianshuai Hu and Xiaolu Liu and Song Wang and Yiyao Zhu and Ao Liang and Lingdong Kong and Guoyang Zhao and Zeying Gong and Jun Cen and Zhiyu Huang and Xiaoshuai Hao and Linfeng Li and Hang Song and Xiangtai Li and Jun Ma and Shaojie Shen and Jianke Zhu and Dacheng Tao and Ziwei Liu and Junwei Liang},
      year={Thu Dec 18 2025 16:57:44 GMT+0000 (Coordinated Universal Time)},
      eprint={2512.16760},
      archivePrefix={arXiv},
      primaryClass={cs.RO},
}

@misc{xu2024drivegpt4interpretableendtoendautonomous,
      title={DriveGPT4: Interpretable End-to-end Autonomous Driving via Large Language Model}, 
      author={Zhenhua Xu and Yujia Zhang and Enze Xie and Zhen Zhao and Yong Guo and Kwan-Yee. K. Wong and Zhenguo Li and Hengshuang Zhao},
      year={2024},
      eprint={2310.01412},
      archivePrefix={arXiv},
      primaryClass={cs.CV},
}

@misc{mao2023gptdriverlearningdrivegpt,
      title={GPT-Driver: Learning to Drive with GPT}, 
      author={Jiageng Mao and Yuxi Qian and Junjie Ye and Hang Zhao and Yue Wang},
      year={2023},
      eprint={2310.01415},
      archivePrefix={arXiv},
      primaryClass={cs.CV},
}

@article{wang2023drivemlm,
  title={DriveMLM: Aligning Multi-Modal Large Language Models with Behavioral Planning States for Autonomous Driving},
  author={Wang, Wenhai and Xie, Jiangwei and Hu, ChuanYang and Zou, Haoming and Fan, Jianan and Tong, Wenwen and Wen, Yang and Wu, Silei and Deng, Hanming and Li, Zhiqi and others},
  journal={arXiv preprint arXiv:2312.09245},
  year={2023}
}

@article{sima2023drivelm,
  title={DriveLM: Driving with Graph Visual Question Answering},
  author={Sima, Chonghao and Renz, Katrin and Chitta, Kashyap and Chen, Li and Zhang, Hanxue and Xie, Chengen and Luo, Ping and Geiger, Andreas and Li, Hongyang},
  journal={arXiv preprint arXiv:2312.14150},
  year={2023}
}

@ARTICLE{chib2024e2eadsurvey,
  author={Chib, Pranav Singh and Singh, Pravendra},
  journal={IEEE Transactions on Intelligent Vehicles}, 
  title={Recent Advancements in End-to-End Autonomous Driving Using Deep Learning: A Survey}, 
  year={2024},
  volume={9},
  number={1},
  pages={103-118},
}

@ARTICLE{le2022survey,
  author={Le Mero, Luc and Yi, Dewei and Dianati, Mehrdad and Mouzakitis, Alexandros},
  journal={IEEE Transactions on Intelligent Transportation Systems}, 
  title={A Survey on Imitation Learning Techniques for End-to-End Autonomous Vehicles}, 
  year={2022},
  volume={23},
  number={9},
  pages={14128-14147},
}

@misc{liang2018cirl,
      title={CIRL: Controllable Imitative Reinforcement Learning for Vision-based Self-driving}, 
      author={Xiaodan Liang and Tairui Wang and Luona Yang and Eric Xing},
      year={2018},
      eprint={1807.03776},
      archivePrefix={arXiv},
      primaryClass={cs.CV},
}

@article{Grigorescu2020survey,
author = {Grigorescu, Sorin and Trasnea, Bogdan and Cocias, Tiberiu and Macesanu, Gigel},
title = {A survey of deep learning techniques for autonomous driving},
journal = {Journal of Field Robotics},
volume = {37},
number = {3},
pages = {362-386},
eprint = {https://onlinelibrary.wiley.com/doi/pdf/10.1002/rob.21918},
year = {2020}
}

@misc{hu2023planningorientedautonomousdriving,
      title={Planning-oriented Autonomous Driving}, 
      author={Yihan Hu and Jiazhi Yang and Li Chen and Keyu Li and Chonghao Sima and Xizhou Zhu and Siqi Chai and Senyao Du and Tianwei Lin and Wenhai Wang and Lewei Lu and Xiaosong Jia and Qiang Liu and Jifeng Dai and Yu Qiao and Hongyang Li},
      year={2023},
      eprint={2212.10156},
      archivePrefix={arXiv},
      primaryClass={cs.CV},
}

@misc{tian2024drivevlm,
      title={DriveVLM: The Convergence of Autonomous Driving and Large Vision-Language Models}, 
      author={Xiaoyu Tian and Junru Gu and Bailin Li and Yicheng Liu and Yang Wang and Zhiyong Zhao and Kun Zhan and Peng Jia and Xianpeng Lang and Hang Zhao},
      year={2024},
      eprint={2402.12289},
      archivePrefix={arXiv},
      primaryClass={cs.CV},
}

@misc{jiang2025diffvla,
      title={DiffVLA: Vision-Language Guided Diffusion Planning for Autonomous Driving}, 
      author={Anqing Jiang and Yu Gao and Zhigang Sun and Yiru Wang and Jijun Wang and Jinghao Chai and Qian Cao and Yuweng Heng and Hao Jiang and Yunda Dong and Zongzheng Zhang and Xianda Guo and Hao Sun and Hao Zhao},
      year={2025},
      eprint={2505.19381},
      archivePrefix={arXiv},
      primaryClass={cs.AI},
}

@misc{pan2024vlp,
      title={VLP: Vision Language Planning for Autonomous Driving}, 
      author={Chenbin Pan and Burhaneddin Yaman and Tommaso Nesti and Abhirup Mallik and Alessandro G Allievi and Senem Velipasalar and Liu Ren},
      year={2024},
      eprint={2401.05577},
      archivePrefix={arXiv},
      primaryClass={cs.CV},
}

@misc{zhou2025autovla,
      title={AutoVLA: A Vision-Language-Action Model for End-to-End Autonomous Driving with Adaptive Reasoning and Reinforcement Fine-Tuning}, 
      author={Zewei Zhou and Tianhui Cai and Seth Z. Zhao and Yun Zhang and Zhiyu Huang and Bolei Zhou and Jiaqi Ma},
      year={2025},
      eprint={2506.13757},
      archivePrefix={arXiv},
      primaryClass={cs.CV},
}

@misc{li2025comprehensivesurveyworldmodels,
      title={A Comprehensive Survey on World Models for Embodied AI}, 
      author={Xinqing Li and Xin He and Le Zhang and Yun Liu},
      year={2025},
      eprint={2510.16732},
      archivePrefix={arXiv},
      primaryClass={cs.CV},
}

@misc{deng2026best3dscenerepresentation,
      title={What Is The Best 3D Scene Representation for Robotics? From Geometric to Foundation Models}, 
      author={Tianchen Deng and Yue Pan and Shenghai Yuan and Dong Li and Chen Wang and Mingrui Li and Long Chen and Lihua Xie and Danwei Wang and Jingchuan Wang and Javier Civera and Hesheng Wang and Weidong Chen},
      year={2026},
      eprint={2512.03422},
      archivePrefix={arXiv},
      primaryClass={cs.RO},
}

@misc{zheng2024gaussianad,
      title={GaussianAD: Gaussian-Centric End-to-End Autonomous Driving}, 
      author={Wenzhao Zheng and Junjie Wu and Yao Zheng and Sicheng Zuo and Zixun Xie and Longchao Yang and Yong Pan and Zhihui Hao and Peng Jia and Xianpeng Lang and Shanghang Zhang},
      year={2024},
      eprint={2412.10371},
      archivePrefix={arXiv},
      primaryClass={cs.CV},
}

@InProceedings{pan2025globalsfmrevisited,
author="Pan, Linfei
and Bar{\'a}th, D{\'a}niel
and Pollefeys, Marc
and Sch{\"o}nberger, Johannes L.",
editor="Leonardis, Ale{\v{s}}
and Ricci, Elisa
and Roth, Stefan
and Russakovsky, Olga
and Sattler, Torsten
and Varol, G{\"u}l",
title="Global Structure-from-Motion Revisited",
booktitle="Computer Vision -- ECCV 2024",
year="2025",
publisher="Springer Nature Switzerland",
address="Cham",
pages="58--77",
isbn="978-3-031-73661-2"
}

@InProceedings{Schonberger2016sfmrevisited,
author = {Schonberger, Johannes L. and Frahm, Jan-Michael},
title = {Structure-From-Motion Revisited},
booktitle = {Proceedings of the IEEE Conference on Computer Vision and Pattern Recognition (CVPR)},
month = {June},
year = {2016}
}

@INPROCEEDINGS{Seitz2006mvs,
  author={Seitz, S.M. and Curless, B. and Diebel, J. and Scharstein, D. and Szeliski, R.},
  booktitle={2006 IEEE Computer Society Conference on Computer Vision and Pattern Recognition (CVPR'06)}, 
  title={A Comparison and Evaluation of Multi-View Stereo Reconstruction Algorithms}, 
  year={2006},
  volume={1},
  number={},
  pages={519-528},
}

@Article{pepe2022uavplatformsandthesfm-mvsapproach,
AUTHOR = {Pepe, Massimiliano and Alfio, Vincenzo Saverio and Costantino, Domenica},
TITLE = {UAV Platforms and the SfM-MVS Approach in the 3D Surveys and Modelling: A Review in the Cultural Heritage Field},
JOURNAL = {Applied Sciences},
VOLUME = {12},
YEAR = {2022},
NUMBER = {24},
ARTICLE-NUMBER = {12886},
ISSN = {2076-3417},
}

@article{mildenhall2021nerf,
  title={Nerf: Representing scenes as neural radiance fields for view synthesis},
  author={Mildenhall, Ben and Srinivasan, Pratul P and Tancik, Matthew and Barron, Jonathan T and Ramamoorthi, Ravi and Ng, Ren},
  journal={Communications of the ACM},
  volume={65},
  number={1},
  pages={99--106},
  year={2021},
  publisher={ACM New York, NY, USA}
}

@misc{xiao2025neuralradiancefieldsreal,
      title={Neural Radiance Fields for the Real World: A Survey}, 
      author={Wenhui Xiao and Remi Chierchia and Rodrigo Santa Cruz and Xuesong Li and David Ahmedt-Aristizabal and Olivier Salvado and Clinton Fookes and Leo Lebrat},
      year={2025},
      eprint={2501.13104},
      archivePrefix={arXiv},
      primaryClass={cs.CV},
}

@article{kerbl20233dgaussian,
  title={3d gaussian splatting for real-time radiance field rendering.},
  author={Kerbl, Bernhard and Kopanas, Georgios and Leimk{\"u}hler, Thomas and Drettakis, George and others},
  journal={ACM Trans. Graph.},
  volume={42},
  number={4},
  pages={139--1},
  year={2023}
}

@misc{huang2024textits3gaussian,
      title={$\textit{S}^3$Gaussian: Self-Supervised Street Gaussians for Autonomous Driving}, 
      author={Nan Huang and Xiaobao Wei and Wenzhao Zheng and Pengju An and Ming Lu and Wei Zhan and Masayoshi Tomizuka and Kurt Keutzer and Shanghang Zhang},
      year={2024},
      eprint={2405.20323},
      archivePrefix={arXiv},
      primaryClass={cs.CV},
}

@ARTICLE{chirstodoulides2025surveyon3drecon,
  author={Christodoulides, Andreas and Tam, Gary K. L. and Clarke, James and Smith, Richard and Horgan, Jon and Micallef, Nicholas and Morley, Jeremy and Villamizar, Nelly and Walton, Sean},
  journal={IEEE Transactions on Visualization and Computer Graphics}, 
  title={Survey on 3D Reconstruction Techniques: Large-Scale Urban City Reconstruction and Requirements}, 
  year={2025},
  volume={31},
  number={10},
  pages={9343-9367},
}

@misc{lin2025depth3recoveringvisual,
      title={Depth Anything 3: Recovering the Visual Space from Any Views}, 
      author={Haotong Lin and Sili Chen and Junhao Liew and Donny Y. Chen and Zhenyu Li and Guang Shi and Jiashi Feng and Bingyi Kang},
      year={2025},
      eprint={2511.10647},
      archivePrefix={arXiv},
      primaryClass={cs.CV},
}

@misc{keetha2026mapanything,
      title={MapAnything: Universal Feed-Forward Metric 3D Reconstruction}, 
      author={Nikhil Keetha and Norman Müller and Johannes Schönberger and Lorenzo Porzi and Yuchen Zhang and Tobias Fischer and Arno Knapitsch and Duncan Zauss and Ethan Weber and Nelson Antunes and Jonathon Luiten and Manuel Lopez-Antequera and Samuel Rota Bulò and Christian Richardt and Deva Ramanan and Sebastian Scherer and Peter Kontschieder},
      year={2026},
      eprint={2509.13414},
      archivePrefix={arXiv},
      primaryClass={cs.CV},
}

@InProceedings{Wang2025vggt,
    author    = {Wang, Jianyuan and Chen, Minghao and Karaev, Nikita and Vedaldi, Andrea and Rupprecht, Christian and Novotny, David},
    title     = {VGGT: Visual Geometry Grounded Transformer},
    booktitle = {Proceedings of the IEEE/CVF Conference on Computer Vision and Pattern Recognition (CVPR)},
    month     = {June},
    year      = {2025},
    pages     = {5294-5306}
}

@misc{wang2025pi3,
      title={$\pi^3$: Permutation-Equivariant Visual Geometry Learning}, 
      author={Yifan Wang and Jianjun Zhou and Haoyi Zhu and Wenzheng Chang and Yang Zhou and Zizun Li and Junyi Chen and Jiangmiao Pang and Chunhua Shen and Tong He},
      year={2025},
      eprint={2507.13347},
      archivePrefix={arXiv},
      primaryClass={cs.CV},
}

@misc{chen2025geodrive,
      title={GeoDrive: 3D Geometry-Informed Driving World Model with Precise Action Control}, 
      author={Anthony Chen and Wenzhao Zheng and Yida Wang and Xueyang Zhang and Kun Zhan and Peng Jia and Kurt Keutzer and Shanghang Zhang},
      year={2025},
      eprint={2505.22421},
      archivePrefix={arXiv},
      primaryClass={cs.CV},
}

@misc{gao2025rad,
      title={RAD: Training an End-to-End Driving Policy via Large-Scale 3DGS-based Reinforcement Learning}, 
      author={Hao Gao and Shaoyu Chen and Bo Jiang and Bencheng Liao and Yiang Shi and Xiaoyang Guo and Yuechuan Pu and Haoran Yin and Xiangyu Li and Xinbang Zhang and Ying Zhang and Wenyu Liu and Qian Zhang and Xinggang Wang},
      year={2025},
      eprint={2502.13144},
      archivePrefix={arXiv},
      primaryClass={cs.CV},
}

@InProceedings{yuan2024presight,
author="Yuan, Tianyuan
and Mao, Yucheng
and Yang, Jiawei
and Liu, Yicheng
and Wang, Yue
and Zhao, Hang",
editor="Leonardis, Ale{\v{s}}
and Ricci, Elisa
and Roth, Stefan
and Russakovsky, Olga
and Sattler, Torsten
and Varol, G{\"u}l",
title="PreSight: Enhancing Autonomous Vehicle Perception with City-Scale NeRF Priors",
booktitle="ECCV",
year="2024",
publisher="Springer Nature Switzerland",
address="Cham",
pages="323--339",
isbn="978-3-031-72980-5"
}

@article{zhao2025pwm,
  title={From Forecasting to Planning: Policy World Model for Collaborative State-Action Prediction},
  author={Zhao, Zhida and Fu, Talas and Wang, Yifan and Wang, Lijun and Lu, Huchuan},
  journal={arXiv preprint arXiv:2510.19654},
  year={2025}
}

@misc{xiong2026unidrivewm,
      title={UniDrive-WM: Unified Understanding, Planning and Generation World Model For Autonomous Driving}, 
      author={Zhexiao Xiong and Xin Ye and Burhan Yaman and Sheng Cheng and Yiren Lu and Jingru Luo and Nathan Jacobs and Liu Ren},
      year={2026},
      eprint={2601.04453},
      archivePrefix={arXiv},
      primaryClass={cs.CV},
}

@article{li2026sgdrive,
  title={SGDrive: Scene-to-Goal Hierarchical World Cognition for Autonomous Driving},
  author={Li, Jingyu and Wu, Junjie and Hu, Dongnan and Huang, Xiangkai and Sun, Bin and Hao, Zhihui and Lang, Xianpeng and Zhu, Xiatian and Zhang, Li},
  journal={arXiv preprint arXiv:2601.05640},
  year={2026}
}

@misc{wu2026infiniteworld,
      title={Infinite-World: Scaling Interactive World Models to 1000-Frame Horizons via Pose-Free Hierarchical Memory}, 
      author={Ruiqi Wu and Xuanhua He and Meng Cheng and Tianyu Yang and Yong Zhang and Zhuoliang Kang and Xunliang Cai and Xiaoming Wei and Chunle Guo and Chongyi Li and Ming-Ming Cheng},
      year={2026},
      eprint={2602.02393},
      archivePrefix={arXiv},
      primaryClass={cs.CV},
}

@misc{zhang2026resworld,
      title={ResWorld: Temporal Residual World Model for End-to-End Autonomous Driving}, 
      author={Jinqing Zhang and Zehua Fu and Zelin Xu and Wenying Dai and Qingjie Liu and Yunhong Wang},
      year={2026},
      eprint={2602.10884},
      archivePrefix={arXiv},
      primaryClass={cs.CV},
}

@inproceedings{daniel2024navsim,
	title = {NAVSIM: Data-Driven Non-Reactive Autonomous Vehicle Simulation and Benchmarking},
	author = {Daniel Dauner and Marcel Hallgarten and Tianyu Li and Xinshuo Weng and Zhiyu Huang and Zetong Yang and Hongyang Li and Igor Gilitschenski and Boris Ivanovic and Marco Pavone and Andreas Geiger and Kashyap Chitta},
	booktitle = {Advances in Neural Information Processing Systems (NeurIPS)},
	year = {2024},
}

@inproceedings{cao2025pseudosimulation, 
	author = {Wei Cao and Marcel Hallgarten and Tianyu Li and Daniel Dauner and Xunjiang Gu and Caojun Wang and Yakov Miron and Marco Aiello and Hongyang Li and Igor Gilitschenski and Boris Ivanovic and Marco Pavone and Andreas Geiger and Kashyap Chitta}, 
	title = {Pseudo-Simulation for Autonomous Driving}, 
	booktitle = {Conference on Robot Learning (CoRL)}, 
	year = {2025}, 
}

@article{xie2024showo,
  title={Show-o: One Single Transformer to Unify Multimodal Understanding and Generation},
  author={Xie, Jinheng and Mao, Weijia and Bai, Zechen and Zhang, David Junhao and Wang, Weihao and Lin, Kevin Qinghong and Gu, Yuchao and Chen, Zhijie and Yang, Zhenheng and Shou, Mike Zheng},
  journal={arXiv preprint arXiv:2408.12528},
  year={2024}
}

@misc{chen2024vadv2,
      title={VADv2: End-to-End Vectorized Autonomous Driving via Probabilistic Planning}, 
      author={Shaoyu Chen and Bo Jiang and Hao Gao and Bencheng Liao and Qing Xu and Qian Zhang and Chang Huang and Wenyu Liu and Xinggang Wang},
      year={2024},
      eprint={2402.13243},
      archivePrefix={arXiv},
      primaryClass={cs.CV},
}

@ARTICLE{chitta2023transfuser,
  author={Chitta, Kashyap and Prakash, Aditya and Jaeger, Bernhard and Yu, Zehao and Renz, Katrin and Geiger, Andreas},
  journal={IEEE Transactions on Pattern Analysis and Machine Intelligence}, 
  title={TransFuser: Imitation With Transformer-Based Sensor Fusion for Autonomous Driving}, 
  year={2023},
  volume={45},
  number={11},
  pages={12878-12895},
}

@misc{zhang2025epona,
      title={Epona: Autoregressive Diffusion World Model for Autonomous Driving}, 
      author={Kaiwen Zhang and Zhenyu Tang and Xiaotao Hu and Xingang Pan and Xiaoyang Guo and Yuan Liu and Jingwei Huang and Li Yuan and Qian Zhang and Xiao-Xiao Long and Xun Cao and Wei Yin},
      year={2025},
      eprint={2506.24113},
      archivePrefix={arXiv},
      primaryClass={cs.CV},
}

@misc{chen2024drivinggpt,
      title={DrivingGPT: Unifying Driving World Modeling and Planning with Multi-modal Autoregressive Transformers}, 
      author={Yuntao Chen and Yuqi Wang and Zhaoxiang Zhang},
      year={2024},
      eprint={2412.18607},
      archivePrefix={arXiv},
      primaryClass={cs.CV},
}

@misc{blattmann2023stablevideodiffusion,
      title={Stable Video Diffusion: Scaling Latent Video Diffusion Models to Large Datasets}, 
      author={Andreas Blattmann and Tim Dockhorn and Sumith Kulal and Daniel Mendelevitch and Maciej Kilian and Dominik Lorenz and Yam Levi and Zion English and Vikram Voleti and Adam Letts and Varun Jampani and Robin Rombach},
      year={2023},
      eprint={2311.15127},
      archivePrefix={arXiv},
      primaryClass={cs.CV},
}

@article{liao2025diffusiondrive,
title={DiffusionDrive: Truncated Diffusion Model for End-to-End Autonomous Driving},
author={Bencheng Liao and Shaoyu Chen and Haoran Yin and Bo Jiang and Cheng Wang and Sixu Yan and Xinbang Zhang and Xiangyu Li and Ying Zhang and Qian Zhang and Xinggang Wang},
booktitle    = {{IEEE/CVF} Conference on Computer Vision and Pattern Recognition,
              {CVPR} 2025, Nashville, TN, USA, June 11-15, 2025},
pages        = {12037--12047},
publisher    = {Computer Vision Foundation / {IEEE}},
year         = {2025},
}

@misc{li2025wote,
      title={End-to-End Driving with Online Trajectory Evaluation via BEV World Model}, 
      author={Yingyan Li and Yuqi Wang and Yang Liu and Jiawei He and Lue Fan and Zhaoxiang Zhang},
      year={2025},
      eprint={2504.01941},
      archivePrefix={arXiv},
      primaryClass={cs.CV},
}

@article{li2025drivevla,
  title={DriveVLA-W0: World Models Amplify Data Scaling Law in Autonomous Driving},
  author={Li, Yingyan and Shang, Shuyao and Liu, Weisong and Zhan, Bing and Wang, Haochen and Wang, Yuqi and Chen, Yuntao and Wang, Xiaoman and An, Yasong and Tang, Chufeng and others},
  journal={arXiv preprint arXiv:2510.12796},
  year={2025}
}

@misc{li2025recogdrive,
      title={ReCogDrive: A Reinforced Cognitive Framework for End-to-End Autonomous Driving}, 
      author={Yongkang Li and Kaixin Xiong and Xiangyu Guo and Fang Li and Sixu Yan and Gangwei Xu and Lijun Zhou and Long Chen and Haiyang Sun and Bing Wang and Kun Ma and Guang Chen and Hangjun Ye and Wenyu Liu and Xinggang Wang},
      year={2025},
      eprint={2506.08052},
      archivePrefix={arXiv},
      primaryClass={cs.CV},
}

@misc{yu2024magvitv2,
      title={Language Model Beats Diffusion -- Tokenizer is Key to Visual Generation}, 
      author={Lijun Yu and José Lezama and Nitesh B. Gundavarapu and Luca Versari and Kihyuk Sohn and David Minnen and Yong Cheng and Vighnesh Birodkar and Agrim Gupta and Xiuye Gu and Alexander G. Hauptmann and Boqing Gong and Ming-Hsuan Yang and Irfan Essa and David A. Ross and Lu Jiang},
      year={2024},
      eprint={2310.05737},
      archivePrefix={arXiv},
      primaryClass={cs.CV},
}

@misc{lu2024wovogen,
      title={WoVoGen: World Volume-aware Diffusion for Controllable Multi-camera Driving Scene Generation}, 
      author={Jiachen Lu and Ze Huang and Zeyu Yang and Jiahui Zhang and Li Zhang},
      year={2024},
      eprint={2312.02934},
      archivePrefix={arXiv},
      primaryClass={cs.CV},
}

@misc{wang2023drivedreamer,
      title={DriveDreamer: Towards Real-world-driven World Models for Autonomous Driving}, 
      author={Xiaofeng Wang and Zheng Zhu and Guan Huang and Xinze Chen and Jiagang Zhu and Jiwen Lu},
      year={2023},
      eprint={2309.09777},
      archivePrefix={arXiv},
      primaryClass={cs.CV},
}

@misc{yang2024genadgeneralizedpredictivemodel,
      title={GenAD: Generalized Predictive Model for Autonomous Driving}, 
      author={Jiazhi Yang and Shenyuan Gao and Yihang Qiu and Li Chen and Tianyu Li and Bo Dai and Kashyap Chitta and Penghao Wu and Jia Zeng and Ping Luo and Jun Zhang and Andreas Geiger and Yu Qiao and Hongyang Li},
      year={2024},
      eprint={2403.09630},
      archivePrefix={arXiv},
      primaryClass={cs.CV},
}

@article{li2025imagidrive,
  title={ImagiDrive: A Unified Imagination-and-Planning Framework for Autonomous Driving},
  author={Li, Jingyu and Zhang, Bozhou and Jin, Xin and Deng, Jiankang and Zhu, Xiatian and Zhang, Li},
  journal={arXiv preprint arXiv:2508.11428},
  year={2025}
}

@article{Li2023phi-1.5,
  title={Textbooks Are All You Need II: \textbf{phi-1.5} technical report},
  author={Li, Yuanzhi and Bubeck, S{\'e}bastien and Eldan, Ronen and Del Giorno, Allie and Gunasekar, Suriya and Lee, Yin Tat},
  journal={arXiv preprint arXiv:2309.05463},
  year={2023}
}

@article{senna,
  title={Senna: Bridging large vision-language models and end-to-end autonomous driving},
  author={Jiang, Bo and Chen, Shaoyu and Liao, Bencheng and Zhang, Xingyu and Yin, Wei and Zhang, Qian and Huang, Chang and Liu, Wenyu and Wang, Xinggang},
  journal={arXiv preprint arXiv:2410.22313},
  year={2024}
}

@inproceedings{vista,
  title={Vista: A generalizable driving world model with high fidelity and versatile controllability},
  author={Gao, Shenyuan and Yang, Jiazhi and Chen, Li and Chitta, Kashyap and Qiu, Yihang and Geiger, Andreas and Zhang, Jun and Li, Hongyang},
  booktitle=NeurIPS,
  year={2024}
}

@article{zhang2025perception,
  title={Perception in Plan: Coupled Perception and Planning for End-to-End Autonomous Driving},
  author={Zhang, Bozhou and Li, Jingyu and Song, Nan and Zhang, Li},
  journal={arXiv preprint arXiv:2508.11488},
  year={2025}
}

@inproceedings{zhangfuture,
  title={Future-Aware End-to-End Driving: Bidirectional Modeling of Trajectory Planning and Scene Evolution},
  author={Zhang, Bozhou and Song, Nan and Zhu, Xiatian and Deng, Jiankang and Zhang, Li and others},
  booktitle={NeurIPS},
  year={2025}
}

@article{zheng2024doe,
  title={Doe-1: Closed-loop autonomous driving with large world model},
  author={Zheng, Wenzhao and Xia, Zetian and Huang, Yuanhui and Zuo, Sicheng and Zhou, Jie and Lu, Jiwen},
  journal={arXiv preprint arXiv:2412.09627},
  year={2024}
}

@article{bartoccioni2025vavim,
  title={Vavim and vavam: Autonomous driving through video generative modeling},
  author={Bartoccioni, Florent and Ramzi, Elias and Besnier, Victor and Venkataramanan, Shashanka and Vu, Tuan-Hung and Xu, Yihong and Chambon, Loick and Gidaris, Spyros and Odabas, Serkan and Hurych, David and others},
  journal={arXiv preprint arXiv:2502.15672},
  year={2025}
}

@article{unterthiner2018fvd,
  title={Towards accurate generative models of video: A new metric \& challenges},
  author={Unterthiner, Thomas and Van Steenkiste, Sjoerd and Kurach, Karol and Marinier, Raphael and Michalski, Marcin and Gelly, Sylvain},
  journal={arXiv preprint arXiv:1812.01717},
  year={2018}
}

@article{caesar2021nuplan,
  title={nuplan: A closed-loop ml-based planning benchmark for autonomous vehicles},
  author={Caesar, Holger and Kabzan, Juraj and Tan, Kok Seang and Fong, Whye Kit and Wolff, Eric and Lang, Alex and Fletcher, Luke and Beijbom, Oscar and Omari, Sammy},
  journal={arXiv preprint arXiv:2106.11810},
  year={2021}
}

\clearpage
\appendix

\section{Unified Discrete Tokenizer}
\label{appendix:tokens}

Our language backbone is Phi-1.5~\cite{Li2023phi-1.5}, 
a text-only large language model that employs a discrete tokenizer with a vocabulary size of 50,295, which cannot directly handle multimodal inputs. 
To enable unified multimodal modeling, we follow the design philosophy of Show-o~\cite{xie2024showo}, 
which represents data from different modalities as discrete tokens within a shared token space. 
This requires extending the original LLM vocabulary with additional visual tokens.

Specifically, images are tokenized by a dual-branch tokenizer consisting of a high‑resolution contextual branch and a low‑resolution dynamic branch, both built upon MagVIT-v2~\cite{yu2024magvitv2} and pretrained with an VQ-GAN  stategy~\cite{zhao2025pwm}, each with an 8192‑entry codebook. To reduce token length, the dynamic branch applies patch-level compression before vector quantization. Through this tokenizer, heterogeneous inputs can be converted into a unified sequence of discrete tokens that can be processed by the LLM. Besides, during inference, the ego status is directly projected into the embedding space using an MLP rather than being discretized into tokens.

\begin{figure}[ht]
  \centering
  \includegraphics[width=0.9\linewidth]{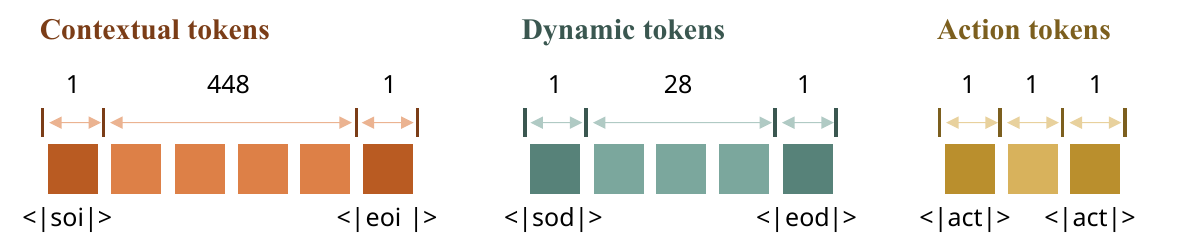}
  \caption{Details of contextual, dynamic and action tokens}
  \label{fig:tokens}
\end{figure}

In addition, we introduce several special tokens, including 
\texttt{<|soi|>}, \texttt{<|eoi|>}, \texttt{<|sod|>}, \texttt{<|eod|>}, 
\texttt{<|t2i|>}, \texttt{<|mmu|>}, \texttt{<|t2d|>}, \texttt{<|act|>}, \texttt{<|sot|>}, \texttt{<|eot|>} and so on. 
The tokens \texttt{<|mmu|>} and \texttt{<|t2d|>} are placed at the beginning of the input 
sequence to indicate the task type. 
\texttt{<|soi|>} and \texttt{<|eoi|>} denote the start and end of contextual image tokens, 
while \texttt{<|sod|>} and \texttt{<|eod|>} mark the boundaries of dynamic image tokens. \texttt{<|sot|>}, \texttt{<|eot|>} indicate the user prompt.
Each trajectory point is enclosed by the token \texttt{<|act|>} to indicate the start and 
end of an action segment, as illustrated in Fig.~\ref{fig:tokens}.

\section{More Visualization}
\label{appendix:vis}

\begin{figure}[p]
  \vfill
  \centering
  \includegraphics[width=\linewidth]{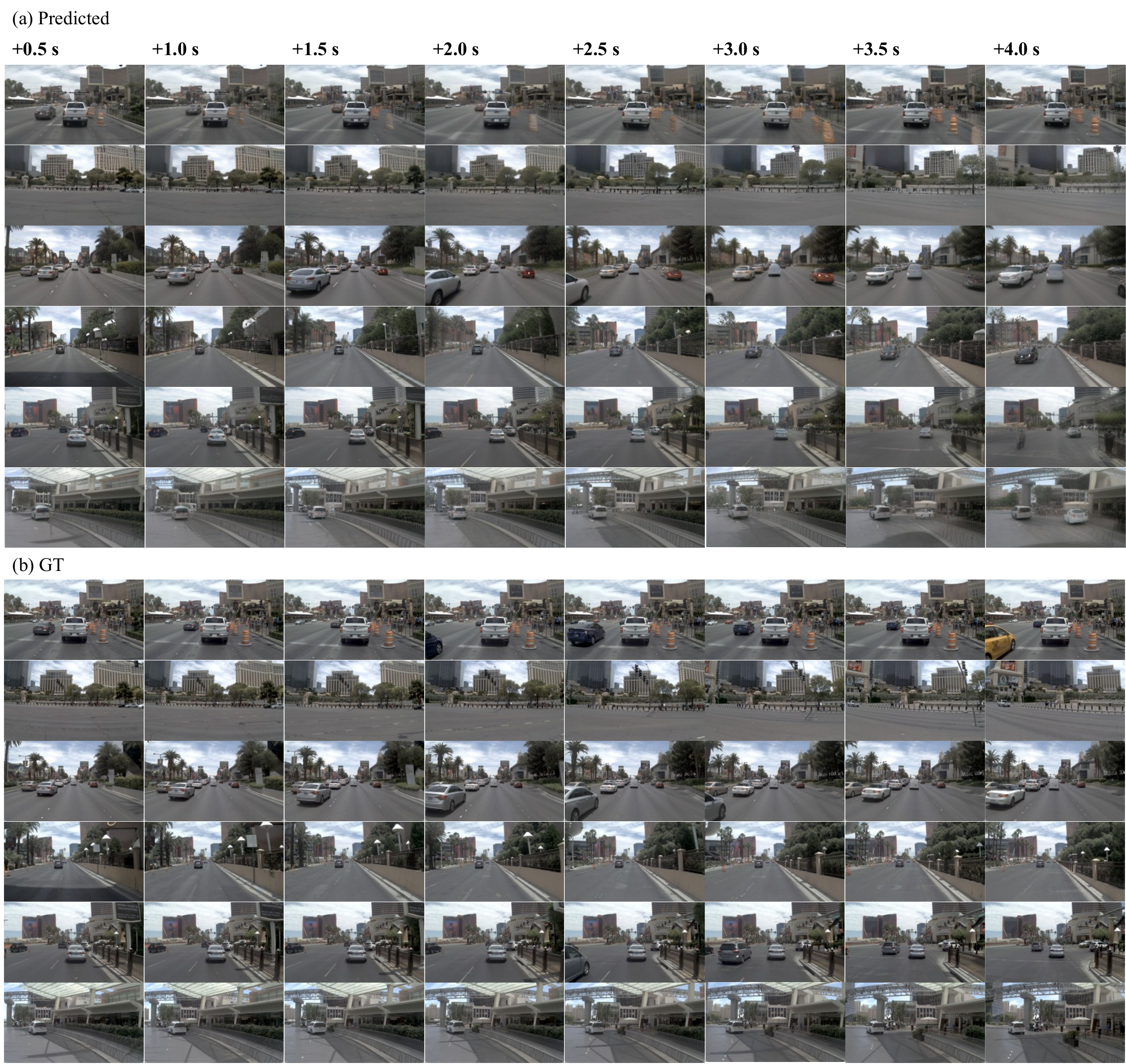}
  \caption{Visualization of predicted future frames (a) and the corresponding ground truth (b) in representative and challenging driving scenarios.}
  \vfill
  \label{fig:more_frames}
\end{figure}

\begin{figure}[ht]
  \centering
  \includegraphics[width=\linewidth]{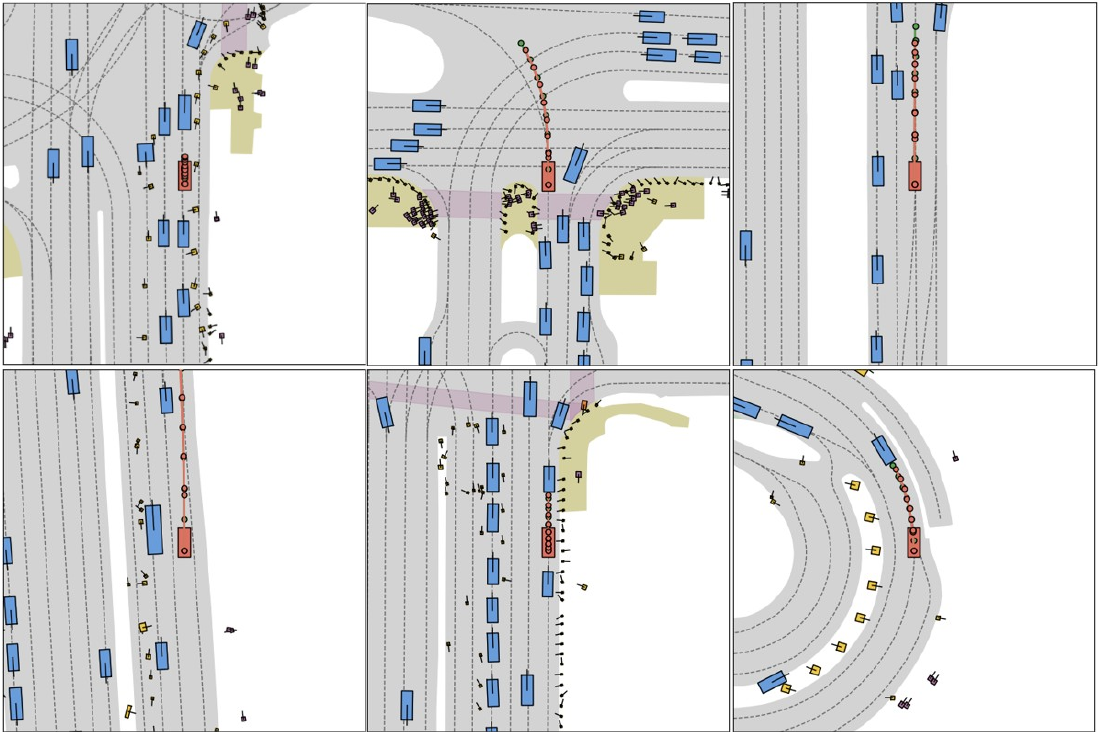}
  \caption{BEV visualization of the corresponding six scenarios (from left to right and top to bottom). 
Green polyline: ground-truth trajectory; red polyline: planned future trajectory.}
  \label{fig:more_bev}
\end{figure}

In this section, we provide additional qualitative visualizations for six representative and challenging driving scenarios. 
Specifically, Fig.~\ref{fig:more_frames} presents the fully decoded predicted future frames together with the corresponding ground-truth frames, illustrating the visual fidelity and temporal consistency of the generated predictions.
Fig.~\ref{fig:more_bev} further shows the BEV visualizations of the same scenarios, highlighting the planned future trajectories compared with the ground-truth trajectories.

\end{document}